\documentclass[12pt,twoside]{article}
\usepackage{latexsym}
\usepackage{wrapfig}

\def\,{\mskip 3mu} \def\>{\mskip 4mu plus 2mu minus 4mu} \def\;{\mskip 5mu plus 5mu} \def\!{\mskip-3mu}
\def\dispmuskip{\thinmuskip= 3mu plus 0mu minus 2mu \medmuskip=  4mu plus 2mu minus 2mu \thickmuskip=5mu plus 5mu minus 2mu}
\def\textmuskip{\thinmuskip= 0mu                    \medmuskip=  1mu plus 1mu minus 1mu \thickmuskip=2mu plus 3mu minus 1mu}
\textmuskip
\def\beq{\dispmuskip\begin{equation}}    \def\eeq{\end{equation}\textmuskip}
\def\beqn{\dispmuskip\begin{displaymath}}\def\eeqn{\end{displaymath}\textmuskip}
\def\bqa{\dispmuskip\begin{eqnarray}}    \def\eqa{\end{eqnarray}\textmuskip}
\def\bqan{\dispmuskip\begin{eqnarray*}}  \def\eqan{\end{eqnarray*}\textmuskip}

\newtheorem{theorem}{Theorem}
\newtheorem{lemma}[theorem]{Lemma}
\newtheorem{definition}[theorem]{Definition}
\newtheorem{figurex}[equation]{Figure}
\newtheorem{corollary}[theorem]{Corollary}

\newenvironment{keywords}{\centerline{\bf\small
Keywords}\vspace{-1ex}\begin{quote}\small}{\par\end{quote}\vskip
1ex}

\def\ftheorem#1#2#3{\begin{theorem}[#2]\label{#1} #3 \end{theorem} }
\def\fcorollary#1#2#3{\begin{corollary}[#2]\label{#1} #3 \end{corollary} }

\def\fdefinition#1#2#3{\begin{definition}[#2]\label{#1} #3 \end{definition} }

\def\paragraph#1{\vspace{1ex}\noindent{\bf{#1.}}}
\def\paranodot#1{\vspace{1ex}\noindent{\bf{#1}}}
\def\toinfty#1{\stackrel{#1\to\infty}{\longrightarrow}}

\def\qed{\hspace*{\fill}$\Box\quad$}
\def\odt{{\textstyle{1\over 2}}}
\def\odf{{\textstyle{1\over 4}}}
\def\eps{\varepsilon}                   
\def\epstr{\epsilon}                    
\def\qmbox#1{{\quad\mbox{#1}\quad}}

\def\equa{\stackrel+=}                 
\def\leqa{\stackrel+\leq}              
\def\geqa{\stackrel+\geq}              
\def\eqm{\stackrel\times=}             
\def\leqm{\stackrel\times\leq}
\def\geqm{\stackrel\times\geq}

\def\l{{l}}                             
\def\M{{\cal M}}                        
\def\X{{\cal X}}                        
\def\Y{{\cal Y}}                        
\def\Q{{\cal Q}}
\def\E{{\bf E}}                         
\def\B{\{0,1\}}                        

\def\MM{M}                              
\def\KM{K\!M}
\def\Km{K\!m}
\def\Set#1{{\if#1Q{I\!\!\!#1}\else\if#1Z{Z\!\!\!Z}\else{I\!\!#1}\fi\fi}}
\def\lb{\log}
\def\mrcp{the chain rule}

\begin{document}

\title{\vskip -5mm\normalsize\sc Technical Report \hfill IDSIA-09-03
\vskip 2mm\bf\LARGE\hrule height5pt \vskip 3mm
\sc Sequence Prediction based \\ on Monotone Complexity%
    \thanks{This work was supported by SNF grant 2000-61847.00 to J\"{u}rgen Schmidhuber.}
\vskip 6mm \hrule height2pt \vskip 5mm}
\author{{\bf Marcus Hutter}\\[3mm]
\normalsize IDSIA, Galleria 2, CH-6928\ Manno-Lugano, Switzerland\\
\normalsize marcus@idsia.ch \hspace{8.5ex} http://www.idsia.ch/$^{_{_\sim}}\!$marcus}
\date{6 June 2003}
\maketitle

\begin{abstract}
\noindent This paper studies sequence prediction based on the
monotone Kolmogorov complexity $\Km=-\lb\,m$, i.e.\ based on
universal deterministic/one-part MDL. $m$ is extremely close to
Solomonoff's prior $M$, the latter being an excellent predictor in
deterministic as well as probabilistic environments, where
performance is measured in terms of convergence of posteriors or
losses. Despite this closeness to $M$, it is difficult to assess
the prediction quality of $m$, since little is known about the
closeness of their posteriors, which are the important quantities
for prediction. We show that for deterministic computable
environments, the ``posterior'' and losses of $m$ converge, but
rapid convergence could only be shown on-sequence; the
off-sequence behavior is unclear. In probabilistic environments,
neither the posterior nor the losses converge, in general.
\end{abstract}


\pagebreak

\section{Introduction}\label{secIntro}

\paragraph{Complexity based sequence prediction}
In this work we study the performance of Occam's razor based
sequence predictors. Given a data sequence $x_1$, $x_2$, ...,
$x_{n-1}$ we want to predict (certain characteristics) of the next
data item $x_n$. Every $x_t$ is an element of some domain $\X$, for
instance weather data or stock-market data at time $t$, or the
$t^{th}$ digit of $\pi$. Occam's razor \cite{Li:97}, appropriately
interpreted, tells us to search for the simplest explanation
(model) of our data $x_1,...,x_{n-1}$ and to use this model for
predicting $x_n$. Simplicity, or more precisely, effective
complexity can be measured by the length of the shortest program
computing sequence $x:=x_1...x_{n-1}$. This length is called the
algorithmic information content of $x$, which we denote by $\tilde
K(x)$. $\tilde K$ stands for one of the many variants of
``Kolmogorov'' complexity (plain, prefix, monotone, ...) or for
$-\lb\,\tilde k(x)$ of universal distributions/measures $\tilde k(x)$.
For simplicity we only consider binary alphabet $\X=\B$ in
this work.

The most well-studied complexity regarding its predictive
properties is $\KM(x)=-\lb M(x)$, where $M(x)$ is Solomonoff's
universal prior \cite{Solomonoff:64,Solomonoff:78}. Solomonoff
has shown that the posterior $M(x_t|x_1...x_{t-1})$ rapidly
converges to the true data generating distribution. In
\cite{Hutter:99errbnd,Hutter:02spupper} it has been shown that $M$
is also an excellent predictor from a decision-theoretic point of
view, where the goal is to minimize loss. In any case, for
prediction, the posterior $M(x_t|x_1...x_{t-1})$, rather than the
prior $M(x_{1:t})$, is the more important quantity.

Most complexities $\tilde K$ coincide within an additive
logarithmic term, which implies that their ``priors'' $\tilde
k=2^{-\tilde K}$ are close within polynomial accuracy. Some of
them are extremely close to each other. Many papers deal with the
proximity of various complexity measures
\cite[...]{Levin:73random,Gacs:83}. Closeness of two complexity
measures is regarded as indication that the quality of their
prediction is similarly good \cite[p.334]{Li:97}. On the other
hand, besides $M$, little is really known about the closeness of
``posteriors'', relevant for prediction.

\paragraph{Aim and conclusion}
The main aim of this work is to study the predictive properties of
complexity measures, other than $\KM$. The monotone complexity
$\Km$ is, in a sense, closest to Solomonoff's complexity $\KM$.
While $\KM$ is defined via a mixture of infinitely many programs,
the conceptually simpler $\Km$ approximates $\KM$ by the
contribution of the single shortest program. This is also closer
to the spirit of Occam's razor. $Km$ is a universal
deterministic/one-part version of the popular Minimal Description
Length (MDL) principle. We mainly concentrate on $\Km$ because it
has a direct interpretation as a universal deterministic/one-part
MDL predictor, and it is closest to the excellent performing
$\KM$, so we expect predictions based on other $\tilde K$ not to
be better.

The main conclusion we will draw is that closeness of priors does
neither necessarily imply closeness of posteriors, nor good
performance from a decision-theoretic perspective. It is far from
obvious, whether $\Km$ is a good predictor in general, and indeed
we show that $\Km$ can fail (with probability strictly greater
than zero) in the presence of noise, as opposed to $\KM$. We do
not suggest that $\Km$ fails for sequences occurring in practice.
It is not implausible that (from a practical point of view) minor
extra (apart from complexity) assumptions on the environment or
loss function are sufficient to prove good performance of $\Km$.
Some complexity measures like $K$, fail completely for prediction.

\paragraph{Contents}
{\em Section \ref{secSetup}} introduces notation and describes how
prediction performance is measured in terms of convergence of
posteriors or losses.
{\em Section \ref{secMProp}} summarizes known predictive
properties of Solomonoff's prior $M$.
{\em Section \ref{secASP}} introduces the monotone complexity
$\Km$ and the prefix complexity $K$ and describes how they and
other complexity measures can be used for prediction.
In {\em Section \ref{secASP}} we enumerate and relate eight
important properties, which general predictive functions may
posses or not: proximity to $M$, universality, monotonicity, being
a semimeasure, \mrcp, enumerability, convergence, and
self-optimizingness. Some later needed normalization issues are
also discussed.
{\em Section \ref{secmProp}} contains our main results. Monotone
complexity $\Km$ is analyzed quantitatively w.r.t.\ the eight
predictive properties. Qualitatively, for deterministic,
computable environments, the posterior converges and is
self-optimizing, but rapid convergence could only be shown
on-sequence; the (for prediction equally important) off-sequence
behavior is unclear. In probabilistic environments, $m$ neither
converges, nor is it self-optimizing, in general.
The proofs are presented in {\em Section \ref{secmProof}}.
{\em Section \ref{secOpen}} contains an outlook and a list of open
questions.

\section{Notation and Setup}\label{secSetup}

\paragraph{Strings and natural numbers}
We write $\X^*$ for the set of finite strings over binary alphabet
$\X=\B$, and $\X^\infty$ for the set of infinity sequences. We use
letters $i,t,n$ for natural numbers, $x,y,z$ for finite strings,
$\epstr$ for the empty string, $\l(x)$ for the length of string
$x$, and $\omega=x_{1:\infty}$ for infinite sequences. We write
$xy$ for the concatenation of string $x$ with $y$.
For a string of length $n$ we write $x_1x_2...x_n$ with
$x_t\in\X$ and further abbreviate $x_{1:n}:=x_1x_2...x_{n-1}x_n$
and $x_{<n}:=x_1... x_{n-1}$. For a given sequence $x_{1:\infty}$
we say that $x_t$ is on-sequence and $\bar x_t\neq x_t$ is
off-sequence. $x'_t$ may be on- or off-sequence.

\paragraph{Prefix sets/codes}
String $x$ is called a (proper) prefix of $y$ if there is a
$z(\neq\epstr)$ such that $xz=y$. We write $x*=y$ in this case,
where $*$ is a wildcard for a string, and similarly for infinite
sequences. A set of strings is called prefix-free if no element is
a proper prefix of another. A prefix-free set $\cal P$ is also
called a prefix-code. Prefix-codes have the important property of
satisfying Kraft's inequality $\sum_{x\in\cal P} 2^{-\l(x)}\leq
1$.

\paragraph{Asymptotic notation}
We abbreviate $\lim_{t\to\infty}[f(t)-g(t)]=0$ by
$f(t)\toinfty{t}g(t)$ and say $f$ converges to $g$, without
implying that $\lim_{t\to\infty}g(t)$ itself exists.
We write $f(x)\leqm  g(x)$ for $f(x)=O(g(x))$ and
$f(x)\leqa  g(x)$ for $f(x)\leq g(x)+O(1)$.
Corresponding equalities can be defined similarly. They hold if
the corresponding inequalities hold in both directions.
$\sum_{t=1}^\infty a_t^2<\infty$ implies $a_t\toinfty{t} 0$. We
say that $a_t$ converges fast or rapidly to zero if
$\sum_{t=1}^\infty a_t^2\leq c$, where $c$ is a constant of
reasonable size; $c=100$ is reasonable, maybe even $c=2^{30}$, but
$c=2^{500}$ is not.$\!$\footnote{Environments of interest have
reasonable complexity $K$, but $2^K$ is not of reasonable size.}
The number of times for which $a_t$ deviates from 0 by more than
$\eps$ is finite and bounded by $c/\eps^2$; no statement is
possible for {\em which} $t$ these deviations occur.
The cardinality of a set $\cal S$ is denoted by $|{\cal S}|$ or
$\#\cal S$.

\paragraph{(Semi)measures}
We call $\rho:\X^*\to[0,1]$ a (semi)measure {\em iff}
$\sum_{x_n\in\X}\rho(x_{1:n}) \stackrel{(<)}= \rho(x_{<n})$ and
$\rho(\epstr) \stackrel{(<)}= 1$. $\rho(x)$ is interpreted as the
$\rho$-probability of sampling a sequence which starts with $x$.
The conditional probability (posterior)
\beq\label{defBayes}
  \rho(x_t|x_{<t}):={\rho(x_{1:t})\over\rho(x_{<t})}
\eeq
is the $\rho$-probability that a string $x_1...x_{t-1}$ is
followed by (continued with) $x_t$.
We call $\rho$ deterministic if
$\exists\omega:\rho(\omega_{1:n})=1$ $\forall n$. In this case we
identify $\rho$ with $\omega$.

\paragraph{Convergent predictors}
We assume that $\mu$ is ``true''\footnote{Also called {\em
objective} or {\em aleatory} probability or {\em chance}.}
sequence generating measure, also called environment. If we know
the generating process $\mu$, and given past data $x_{<t}$ we can
predict the probability $\mu(x_t|x_{<t})$ of the next data item
$x_t$. Usually we do not know $\mu$, but estimate it from
$x_{<t}$. Let $\rho(x_t|x_{<t})$ be an estimated
probability\footnote{Also called {\em subjective} or {\em belief}
or {\em epistemic} probability.} of $x_t$, given $x_{<t}$.
Closeness of $\rho(x_t|x_{<t})$ to $\mu(x_t|x_{<t})$ is expected
to lead to ``good'' predictions:

Consider, for instance, a weather data sequence $x_{1:n}$ with
$x_t=1$ meaning rain and $x_t=0$ meaning sun at day $t$. Given
$x_{<t}$ the probability of rain tomorrow is $\mu(1|x_{<t})$. A
weather forecaster may announce the probability of rain to be
$y_t:=\rho(1|x_{<t})$, which should be close to
the true probability $\mu(1|x_{<t})$.
To aim for
\beq\label{eqconv}
 \rho(x'_t|x_{<t}) \;\stackrel{(fast)}\longrightarrow\; \mu(x'_t|x_{<t})
 \qmbox{for} t\to\infty
\eeq
seems reasonable. A sequence of random variables $z_t=z_t(\omega)$
(like $z_t=\rho(x_t|x_{<t})-\mu(x_t|x_{<t})$) is said to converge
to zero with $\mu$-probability 1 (w.p.1) if the set $\{\omega :
z_t(\omega)\toinfty{t} 0\}$ has $\mu$-measure 1. $z_t$ is said to
converge to zero in mean sum (i.m.s) if
$\sum_{t=1}^\infty\E[z_t^2]\leq c<\infty$, where $\E$ denotes
$\mu$-expectation. Convergence i.m.s.\ implies convergence w.p.1
(rapid if $c$ is of reasonable size).

Depending on the interpretation, a $\rho$ satisfying
(\ref{eqconv}) could be called consistent or self-tuning
\cite{Kumar:86}. One problem with using (\ref{eqconv}) as
performance measure is that closeness cannot be computed, since
$\mu$ is unknown. Another disadvantage is that (\ref{eqconv}) does
not take into account the value of correct predictions or the
severity of wrong predictions.

\paragraph{Self-optimizing predictors}
More practical and flexible is a decision-theoretic approach,
where performance is measured w.r.t.\ the true outcome sequence
$x_{1:n}$ by means of a loss function, for instance
$\ell_{x_ty_t}:=(x_t-y_t)^2$, which does not involve $\mu$.
More generally, let $\ell_{x_t y_t}\in[0,1]\subset\Set{R}$ be the
received loss when performing some prediction/decision/action
$y_t\in\Y$ and $x_t\in\X$ is the $t^{th}$ symbol of the sequence.
Let $y_t^\Lambda \in \Y$ be the prediction of a (causal)
prediction scheme $\Lambda$. The true probability of the next
symbol being $x_t$, given $x_{<t}$, is $\mu(x_t|x_{<t})$. The
$\mu$-expected loss (given $x_{<t}$) when $\Lambda$ predicts the
$t^{th}$ symbol is
\beqn\label{loss2}
  l_t^\Lambda(x_{<t}) \;:=\;
  \sum_{x_t}\mu(x_t|x_{<t}) \ell_{x_t y_t^\Lambda}.
\eeqn
The goal is
to minimize the $\mu$-expected loss. More generally, we define the
$\Lambda_\rho$ sequence prediction scheme
\beq\label{xlrdef}
  y_t^{\Lambda_\rho} :=
  \arg\min_{y_t\in\Y}\sum_{x_t}\rho(x_t|x_{<t})\ell_{x_t y_t},
\eeq
which minimizes the $\rho$-expected loss. If $\mu$ is known,
$\Lambda_\mu$ is obviously the best prediction scheme in the sense
of achieving minimal expected loss ($l_t^{\Lambda_\mu}\leq
l_t^{\Lambda}$ for all $\Lambda$). An important special case is
the error-loss $\ell_{xy}=1-\delta_{xy}$ with $\Y=\X$. In this
case $\Lambda_\rho$ predicts the $y_t$ which maximizes
$\rho(y_t|x_{<t})$,  and $\sum_t\E[l_t^{\Lambda_\rho}]$ is the
expected number of prediction errors (where
$y_t^{\Lambda_\rho}\neq x_t$).
The natural decision-theoretic counterpart of (\ref{eqconv}) is to
aim for
\beq\label{eqlconv}
 l_t^{\Lambda_\rho}(x_{<t}) \;\stackrel{(fast)}\longrightarrow\;
 l_t^{\Lambda_\mu}(x_{<t}) \qmbox{for} t\to\infty
\eeq
what is called (without the fast supplement) self-optimizingness
in control-theory \cite{Kumar:86}.

\section{Predictive Properties of $M=2^{-\KM}$}\label{secMProp}

We define a prefix Turing machine $T$ as a Turing machine with
binary unidirectional input and output tapes, and some
bidirectional work tapes. We say $T$ halts on input $p$ with
output $x$ and write ``$T(p)=x$ halts'' if $p$ is to the left of
the input head and $x$ is to the left of the output head after $T$
halts. The set of $p$ on which $T$ halts forms a prefix-code. We
call such codes $p$ {\em self-delimiting} programs. We write
$T(p)=x*$ if $T$ outputs a string starting with $x$; $T$ need not
to halt in this case. $p$ is called {\em minimal} if $T(q)\neq x*$
for all proper prefixes of $p$. The set of all prefix
Turing-machines $\{T_1,T_2,...\}$ can be effectively enumerated.
There exists a universal prefix Turing machine $U$ which can
simulate every $T_i$. A function is called computable if there is
a Turing machine, which computes it. A function is called
enumerable if it can be approximated from below. Let
$\M_{comp}^{msr}$ be the set of all computable measures,
$\M_{enum}^{semi}$ the set of all enumerable semimeasures, and
$\M_{det}$ be the set of all deterministic measures
($\widehat=\X^\infty$).$\!$\footnote{$\M_{enum}^{semi}$ is
enumerable, but $\M_{comp}^{msr}$ is not, and $\M_{det}$ is
uncountable.}

Levin \cite{Zvonkin:70,Li:97} has shown the existence of an
enumerable universal semimeasure $\MM$
($\MM\geqm\nu$ $\forall\nu\in\M_{enum}^{semi}$). An
explicit expression due to Solomonoff \cite{Solomonoff:78} is
\beq\label{defM}\label{defKM}
  \MM(x) \;:=\; \sum_{p:U(p)=x*}2^{-\l(p)}, \qquad
  \KM(x) := -\lb M(x).
\eeq
The sum is over all (possibly non-halting) minimal programs $p$
which output a string starting with $x$. This definition is
equivalent to the probability that $U$ outputs a string starting
with $x$ if provided with fair coin flips on the input tape. $M$
can be used to characterize randomness of individual sequences: A
sequence $x_{1:\infty}$ is (Martin-L\"{o}f) $\mu$-random, {\em
iff} $\exists c:M(x_{1:n})\leq c\cdot\mu(x_{1:n})\forall n$.
For later comparison, we summarize the (excellent) predictive
properties of $\MM$
\cite{Solomonoff:78,Hutter:01alpha,Hutter:02spupper} (the
numbering will become clearer later):

\ftheorem{thMProp}{Properties of $\MM=2^{-\KM}$}{
Solomonoff's prior $\MM$ defined in (\ref{defM}) is a $(i)$
universal, $(v)$ enumerable, $(ii)$ monotone, $(iii)$ semimeasure,
which $(vi)$ converges to $\mu$ i.m.s., and $(vii)$ is
self-optimizing i.m.s. More quantitatively:
\begin{itemize}\parskip=0ex\parsep=0ex\itemsep=1ex
\item[$(vi)$]
  $\sum_{t=1}^\infty\E[\sum_{x'_t}(M(x'_t|x_{<t})-\mu(x'_t|x_{<t}))^2]
  \;\leqa\; \ln 2\cdot K(\mu)$, which implies \\
  $M(x'_t|x_{<t}) \;\toinfty{t}\; \mu(x'_t|x_{<t})$ i.m.s.\ for
  $\mu\in\M_{comp}^{msr}$.
\item[$(vii)$]
  $\sum_{t=1}^\infty\E[(l_t^{\Lambda_M}-l_t^{\Lambda_\mu})^2]
  \;\leqa\;  2\ln 2\cdot K(\mu)$, which implies \\
  $l_t^{\Lambda_M} \;\toinfty{t}\; l_t^{\Lambda_\mu}$ i.m.s.\ for
  $\mu\in\M_{comp}^{msr}$,
\end{itemize}
where $K(\mu)$ is the length of the shortest program computing
function $\mu$.
}

\section{Alternatives to Solomonoff's Prior $\MM$}\label{secASP}

The goal of this work is to investigate whether some other
quantities which are closely related to $\MM$ also lead to good
predictors. The prefix Kolmogorov complexity $K$
is closely
related to $\KM$ ($K(x)=\KM(x)+O(\log\,\l(x))$).
$K(x)$ is defined as the
length of the shortest halting program on $U$ with output $x$:
\beq\label{defK}\label{defk}
  K(x) := \min\{\l(p): U(p)=x \mbox{ halts} \}, \qquad
  k(x):=2^{-K(x)}.
\eeq
In Section \ref{secOpen} we briefly
discuss that $K$ completely fails for predictive purposes.
More promising is to approximate
$\MM(x)=\sum_{p:U(p)=x*}2^{-\l(p)}$ by the dominant contribution
in the sum, which is given by
\beq\label{defm}\label{defKm}
  m(x):=2^{-\Km(x)} \qmbox{with}
  \Km(x):=\min_p\{\l(p):U(p)=x*\}.
\eeq
$\Km$ is called {\em monotone complexity} and has been shown to be
{\em very} close to $\KM$ \cite{Levin:73random,Gacs:83} (see also Theorem
\ref{thmDProp}$(o)$). It is natural to call a sequence
$x_{1:\infty}$ {\em computable} if $\Km(x_{1:\infty})<\infty$.
$\KM$, $\Km$, and $K$ are ordered in the following way:
\beq\label{Krels}
  0 \;\leq\; K(x|\l(x)) \;\leqa\; \KM(x) \;\leq\; \Km(x) \;\leq\; K(x)
  \;\leqa\; \l(x) + 2\lb\l(x).
\eeq
There are many complexity measures (prefix, Solomonoff, monotone,
plain, process, extension, ...) which we generically denote by
$\tilde K\in\{K,\KM,\Km,...\}$ and their associated ``predictive
functions'' $\tilde k(x):=2^{-\tilde K(x)}\in\{k,M,m,...\}$. This
work is mainly devoted to the study of $m$.

Note that $\tilde k$ is generally not a semimeasure, so we have to
clarify what it means to predict using $\tilde k$. One popular
approach which is at the heart of the (one-part) MDL principle is
to predict the $y$ which minimizes $\tilde K(xy)$ (maximizes
$\tilde k(xy))$, where $x$ are past given data:
$y_t^{MDL}:=\arg\min_{y_t}\tilde K(x_{<t}y_t)$.

For complexity measures $\tilde K$, the conditional version
$\tilde K_|(x|y)$ is often defined\footnote{Usually written
without index $|$.} as $\tilde K(x)$, but where the underlying
Turing machine $U$ has additionally access to $y$.
The definition $\tilde k_|(x|y):=2^{-\tilde K_|(x|y)}$ for the
conditional predictive function $\tilde k$ seems natural, but has
the disadvantage that the crucial \mrcp\ (\ref{defBayes}) is
violated. For $\tilde K=K$ and $\tilde K=\Km$ and most other
versions of $\tilde K$, \mrcp\ is still satisfied
approximately (to logarithmic accuracy), but this is not
sufficient to prove convergence (\ref{eqconv}) or
self-optimizingness (\ref{eqlconv}). Therefore, we define $\tilde
k(x_t|x_{<t}):=\tilde k(x_{1:t})/\tilde k(x_{<t})$ in the
following, analogously to semimeasures $\rho$ (like $\MM$). A
potential disadvantage of this definition is that $\tilde
k(x_t|x_{<t})$ is not enumerable, whereas $\tilde k_|(x_t|x_{<t})$
and $\tilde k(x_{1:t})$ are.

We can now embed MDL predictions minimizing $\tilde K$ into our
general framework: MDL coincides with the $\Lambda_{\tilde k}$
predictor for the error-loss:
\beq\label{eqMDLk}
  y_t^{\Lambda_{\tilde k}} \;=\;
  \arg\max_{y_t}\tilde k(y_t|x_{<t}) \;=\;
  \arg\max_{y_t}\tilde k(x_{<t}y_t) \;=\;
  \arg\min_{y_t}\tilde K(x_{<t}y_t) \;=\;
  y_t^{MDL}
\eeq
In the first equality we inserted $\ell_{xy}=1-\delta_{xy}$ into
(\ref{xlrdef}). In the second equality we used \mrcp\
(\ref{defBayes}). In both steps we dropped some in $\arg\max$
ineffective additive/multiplicative terms independent of $y_t$. In
the third equality we used $\tilde k=2^{-\tilde K}$. The last
equality formalizes the one-part MDL principle: given $x_{<t}$
predict the $y_t\in\X$ which leads to the shortest code $p$.
Hence, validity of (\ref{eqlconv}) tells us something about the
validity of the MDL principle. (\ref{eqconv}) and (\ref{eqlconv})
address what (good) prediction {\em means}.

\section{General Predictive Functions}\label{secGPF}

We have seen that there are predictors (actually the major one
studied in this work) $\Lambda_\rho$, but where $\rho(x_t|x_{<t})$
is not (immediately) a semimeasure. Nothing prevents us from
replacing $\rho$ in (\ref{xlrdef}) by an arbitrary function
$b_|:\X^*\to[0,\infty)$, written as $b_|(x_t|x_{<t})$.
We also define general functions $b:\X^*\to[0,\infty)$, written as
$b(x_{1:n})$ and $b(x_t|x_{<t}):={b(x_{1:t})\over b(x_{<t})}$,
which may not coincide with $b_|(x_t|x_{<t})$. Most terminology for
semimeasure $\rho$ can and will be carried over to the case of
general predictive functions $b$ and $b_|$, but one has to be
careful which properties and interpretations still hold:

\fdefinition{defbProp}{Properties of predictive functions}{
We call functions $b,b_|:\X^*\to[0,\infty)$ (conditional)
predictive functions. They may possess some of the following
properties:
\begin{itemize}\parskip=0ex\parsep=0ex\itemsep=0.5ex\itemindent1ex
\item[$o)$] {\em Proximity:} $b(x)$ is ``close'' to the universal
prior $\MM(x)$
\item[$i)$] {\em Universality:} $b\geqm\M$, i.e.\
$\forall\nu\in\M\,\exists c>0: b(x)\geq c\cdot\nu(x)\forall x$.
\item[$ii)$] {\em Monotonicity:} $b(x_{1:t})\leq
b(x_{<t})\;\forall t,x_{1:t}$
\item[$iii)$] {\em Semimeasure:} $\sum_{x_t}b(x_{1:t})\leq
b(x_{<t})$ and $b(\epstr)\leq 1$
\item[$iv)$] {\em Chain rule:} $b(x_{1:t})=b.(x_t|x_{<t})b(x_{<t})$
\item[$v)$] {\em Enumerability:} $b$ is lower semi-computable
\item[$vi)$] {\em Convergence:}
$b.(x'_t|x_{<t})\toinfty{t}\mu(x'_t|x_{<t})$ $\forall \mu\in\M,
x'_t\in\X$ i.m.s.\ or w.p.1
\item[$vii)$] {\em Self-optimizingness:}
$l_t^{\Lambda_{b.}} \;\toinfty{t}\; l_t^{\Lambda_\mu}$ i.m.s.\ or w.p.1
\end{itemize}
where $b.$ refers to $b$ or $b_|$
}

\noindent The importance of the properties $(i)-(iv)$ stems from
the fact that they together imply convergence $(vi)$ and
self-optimizingness $(vii)$. Regarding proximity $(o)$ we left
open what we mean by ``close''. We also did not specify $\M$ but
have in mind all computable measures $\M_{comp}^{msr}$ or
enumerable semimeasures $\M_{enum}^{semi}$, possibly restricted to
deterministic environments $\M_{det}$.

\ftheorem{thPredRel}{Predictive relations}{\hfill
\begin{itemize}\parskip=0ex\parsep=0ex\itemsep=1ex
\item[$a)$] $(iii)\Rightarrow(ii)$: A semimeasure is monotone.
\item[$b)$] $(i),(iii),(iv)\Rightarrow(vi)$: The posterior $b.$ as defined
by \mrcp\ $(iv)$ of a universal semimeasure $b$ converges to
$\mu$ i.m.s.\ for all $\mu\in\M$.
\item[$c)$] $(i),(iii),(v)\Rightarrow(o)$: Every w.r.t.\ $\M_{enum}^{semi}$
universal enumerable semimeasure coincides with $M$ within a multiplicative constant.
\item[$d)$] $(vi)\Rightarrow(vii)$: Posterior convergence i.m.s./w.p.1
implies self-optimizingness i.m.s./w.p.1.
\end{itemize}
}

\paragraph{Proof sketch} %
$(a)$ follows trivially from dropping the sum in $(ii)$, %
$(b)$ and $(c)$ are Solomonoff's major results
\cite{Solomonoff:78,Li:97,Hutter:01alpha}, %
$(d)$ follows from $0\leq l_t^{\Lambda_{b.}}-l_t^{\Lambda_\mu}
\leq\sum_{x'_t}|b.(x'_t|x_{<t})-\mu(x'_t|x_{<t})|$, since
$\ell\in[0,1]$ \cite[Th.4$(ii)$]{Hutter:02spupper}.
\qed

We will see that $(i),(iii),(iv)$ are crucial for proving
$(vi),(vii)$.

\paragraph{Normalization}
Let us consider a scaled $b$ version
$b_{norm}(x_t|x_{<t}):=c(x_{<t})b(x_t|x_{<t})$, where $c>0$ is
independent of $x_t$. Such a scaling does not affect the
prediction scheme $\Lambda_b$ (\ref{xlrdef}), i.e.\
$y_t^{\Lambda_b}=y_t^{\Lambda_{b_{norm}}}$, which implies
$l_t^{\Lambda_{b_{norm}}}=l_t^{\Lambda_b}$. Convergence
$b(x'_t|x_{<t})\to\mu(x'_t|x_{<t})$ implies
$\sum_{x'_t}b(x'_t|x_{<t})\to 1$ if $\mu$ is a measure, hence also
$b_{norm}(x'_t|x_{<t})\to\mu(x'_t|x_{<t})$
for\footnote{Arbitrarily we define
$b_{norm}(x_t|x_{<t})={1\over|\X|}$ if
$\sum_{x'_t}b(x'_t|x_{<t})=0$.}
$c(x_{<t}):=[\sum_{x'_t}b(x'_t|x_{<t})]^{-1}$. Speed of
convergence may be affected by normalization, either positively or
negatively. Assuming \mrcp\ (\ref{defBayes}) for $b_{norm}$ we get
\beqn
  b_{norm}(x_{1:n}) =
  \prod_{t=1}^n{b(x_{1:t})\over\sum_{x_t}b(x_{1:t})} =
  d(x_{<n})b(x_{1:n}), \qquad
  d(x_{<n}):={1\over b(\epstr)}\prod_{t=1}^n{b(x_{<t})\over\sum_{x_t}b(x_{1:t})}
\eeqn
Whatever $b$ we start with, $b_{norm}$ is a measure, i.e.\ $(iii)$
is satisfied with equality. Convergence and self-optimizingness
proofs are now eligible for $b_{norm}$, provided universality $(i)$
can be proven for $b_{norm}$.
If $b$ is a semimeasure, then $d\geq 1$, hence
$\MM_{norm}\geq\MM\geqm\M_{enum}^{semi}$ is
universal and converges $(vi)$ with same bound (Theorem
\ref{thMProp}$(vi)$) as for $\MM$. On the other hand $d(x_{<n})$
may be unbounded for $b=k$ and $b=m$, so
normalization does not help us in these cases for proving $(vi)$.
Normalization transforms a universal non-semimeasure into a
measure, which may no longer be universal.

\section{Predictive Properties of $m=2^{-\Km}$}\label{secmProp}

We can now state which predictive properties of $m$ hold, and
which not. In order not to overload the reader, we first summarize
the qualitative predictive properties of $m$ in Corollary
\ref{thmProp}, and subsequently present detailed quantitative
results in Theorem \ref{thmDProp}, followed by an item-by-item
explanation and discussion. The proofs are deferred to the next
section.

\fcorollary{thmProp}{Properties of $m=2^{-\Km}$}{
For $b=m=2^{-\Km}$, where $\Km$ is the monotone Kolmogorov
complexity (\ref{defKm}), the following properties of Definition
\ref{defbProp} are satis\-fied\-/\-vio\-lated: %
$(o)$ For every $\mu\in\M_{comp}^{msr}$ and every $\mu$-random
sequence $x_{1:\infty}$, $m(x_{1:n})$ equals $M(x_{1:n})$ within a
multiplicative constant. %
$m$ is $(i)$ universal (w.r.t.\ $\M=\M_{comp}^{msr}$), %
$(ii)$ monotone, %
and $(v)$ enumerable, %
but is $\neg(iii)$ not a semimeasure. %
$m$ satisfies $(iv)$ \mrcp\ by definition for
$m.=m$, but for $m.=m_|$ \mrcp\ is only satisfied to logarithmic order. %
For $m.=m$, $m$ $(vi)$ converges and $(vii)$ is self-optimizing
for deterministic $\mu\in\M_{comp}^{msr}\cap\M_{det}$, but in
general not for probabilistic $\mu\in\M_{comp}^{msr}\setminus\M_{det}$. %
}

\noindent The lesson to learn is that although $m$ is very close
to $M$ in the sense of $(o)$ and $m$ dominates all computable
measures $\mu$, predictions based on $m$ may nevertheless fail
(cf. Theorem \ref{thMProp}).

\ftheorem{thmDProp}{Detailed properties of $m=2^{-\Km}$}{
For $b=m=2^{-\Km}$, where $\Km(x):=\min_p\{\l(p):U(p)=x*\}$ is the
monotone Kolmogorov complexity, the following properties of
Definition \ref{defbProp} are satisfied / violated:
\begin{itemize}\parskip=0ex\parsep=0ex\itemindent0ex\itemsep=1ex
\item[$(o)$]
$\forall\mu\in\M_{comp}^{msr}\,\forall\mu$-random $\omega\,\exists
c_\omega \,:\, \Km(\omega_{1:n})\leq \KM(\omega_{1:n})+c_\omega\,\forall
n$, \hfill \cite{Levin:73random} \\%
$\KM(x)\leq \Km(x)\leq \KM(x)+2\,\lb \Km(x)+O(1)\,\forall x$. \hfill \cite[Th.3.4]{Zvonkin:70}
\item[$\neg(o)$]
$\forall c \,:\, \Km(x)-\KM(x)\geq c$ for infinitely many $x$. \hfill \cite{Gacs:83}
\item[$(i)$] $\Km(x)\leqa -\lb\,\mu(x)+K(\mu)$ if
$\mu\in\M_{comp}^{msr}$, \hfill \cite[Th.4.5.4]{Li:97} \\[-0.5ex] 
$m\geqm\M_{comp}^{msr}$, but
$m\not\geqm\M_{enum}^{semi}$ (unlike
$\MM\geqm\M_{enum}^{semi}$).
\item[$(ii)$] $\Km(xy)\geq \Km(x)\in\Set N_0$,
$\quad 0<m(xy)\leq m(x)\in 2^{-\Set N_0}\leq 1$.
\item[$\neg(iii)$]
If $x_{1:\infty}$ is computable, then $\sum_{x_t}m(x_{1:t})\not\leq m(x_{<t})$ for almost all $t$, \\
If $\;\Km(x_{1:t})=o(t)$, $\;$ then $\sum_{x_t}m(x_{1:t})\not\leq m(x_{<t})$ for most $t$.
\item[$(iv)$]
$0 < m(x|y):= {m(yx)\over m(y)} \leq 1$.
\item[$\neg(iv)$]
if $m_|(x|y):= 2^{-\min_p\{\l(p):U(p,y)=x*\} }$, then
$\exists x,y : m(yx)\neq m_|(x|y)\cdot m(y)$, \\
$\Km(yx)=\Km_|(x|y)+\Km(y) \pm O(\log\,\l(xy))$.
\item[$(v)$] $m$ is enumerable, i.e.\ lower semi-computable.
\item[$(vi)$] $\sum_{t=1}^n|1-m(x_t|x_{<t})|\leq \odt
\Km(x_{1:n})$, $\quad
m(x_t|x_{<t})\stackrel{fast}\longrightarrow 1$ for comp.\
$x_{1:\infty}$, \\
Indeed, $m(x_t|x_{<t})\neq 1$ at most $\Km(x_{1:\infty})$ times, \\
$\sum_{t=1}^n
m(\bar x_t|x_{<t}) \leq 2^{\Km(x_{1:n})}$, $\quad
m(\bar x_t|x_{<t})\stackrel{slow?}\longrightarrow 0$ for
computable $x_{1:\infty}$.
\item[$\neg(vi)$]
$\exists\mu\in\M_{comp}^{msr}\setminus\M_{det} \,:\,
m_{(norm)}(x_t|x_{<t}) \not\to \mu(x_t|x_{<t})\,\forall x_{1:\infty}$
\item[$(vii)$]
$l_t^{\Lambda_m}(x_{<t}) \;\stackrel{slow?}\longrightarrow\;
l_t^{\Lambda_\omega}:=\arg\min_{y_t}\ell_{x_ty_t}$
if $\omega\equiv x_{1:\infty}$ is computable. \\
$\Lambda_m=\Lambda_{m_{norm}}$, i.e.\
$y_t^{\Lambda_m}=y_t^{\Lambda_{m_{norm}}}$ and
$l_t^{\Lambda_m}=l_t^{\Lambda_{m_{norm}}}$.
\item[$\neg(vii)$] $\forall |\Y|>2\,\exists\ell,
\mu \;:\; {l_t^{\Lambda_m}/l_t^{\Lambda_\mu}}
= c > 1\,\forall t$
$\quad (c={6\over 5}-\eps$ possible$)$.
\\
$\forall$ non-degenerate $\ell\;\exists U,\mu \;:\;
{l_t^{\Lambda_m}/l_t^{\Lambda_\mu}} \;\;\not\!\!\!\toinfty{t} 1$
with high probability.
\end{itemize}
}

\paragraph{Explanation and discussion}
$(o)$ The first line shows that $m$ is close to $M$
within a multiplicative constant for nearly all strings in a very
strong sense. $\sup_n{M(\omega_{1:n})\over m(\omega_{1:n})}\leq
2^{c_\omega}$ is finite for every $\omega$ which is random (in
the sense of Martin-L{\"o}f) w.r.t.\ {\em any} computable $\mu$,
but note that the constant $c_\omega$ depends on $\omega$. Levin
falsely conjectured the result to be true for {\em all} $\omega$,
but could only prove it to hold within logarithmic accuracy
(second line).

$\neg(o)$ A later result by G{\'a}cs, indeed, implies that
$\Km-\KM$ is unbounded (for infinite alphabet it can even increase
logarithmically).

$(i)$ The first line can be interpreted as a ``continuous'' coding
theorem for $\Km$ and recursive $\mu$. It implies (by
exponentiation) that $m$ dominates all computable measures (second
line). Unlike $\MM$ it does {\em not} dominate all enumerable
semimeasures. Dominance is a key feature for good predictors. From
a practical point of view the assumption that the true generating
distribution $\mu$ is a proper measure and computable seems not to
be restrictive. The problem will be that $m$ is not a semimeasure.

$(ii)$ The monotonicity property is obvious from the definition of
$\Km$ and is the origin of calling $\Km$ monotone complexity.

$\neg(iii)$ shows and quantifies how the crucial semimeasure
property is violated for $m$ in an essential way, where {\em
almost all} $n$ means ``all but finitely many,$\!$'' and {\em most} $n$
means ``all but an asymptotically vanishing fraction.$\!$''.

$(iv)$ \mrcp\ can be satisfied by definition. With such a
definition, $m(x|y)$ is strictly positive like $M(x|y)$, but not
necessarily strictly less than $1$, unlike $M(x|y)$. Nevertheless
it is bounded by $1$ due to monotonicity of $m$, unlike for $k$.

$\neg(iv)$ If a conditional monotone complexity $\Km_|=-\lb\,m_|$
is defined similarly to the conditional Kolmogorov complexity $K_|$,
then \mrcp\ is only valid within logarithmic accuracy.

$(v)$ $m$ shares the obvious enumerability property with $M$.

$(vi)$ (first line) shows that the on-sequence predictive
properties of $m$ for deterministic computable environments are
excellent. The predicted $m$-probability\footnote{We say
``probability'' just for convenience, not forgetting that
$m(\cdot|x_{<t})$ is not a proper (semi)probability distribution.}
of $x_t$ given $x_{<t}$ converges rapidly to 1 for reasonably
simple/complex $x_{1:\infty}$. A similar result holds for $M$. The
stronger result (second line), that $m(x_t|x_{<t})$ deviates from
1 at most $\Km(x_{1:\infty})$ times, does not hold for $M$.
Note that perfect on-sequence prediction could trivially be
achieved by always predicting 1 ($b.\equiv 1$).
Since we do not know the true outcome $x_t$ in advance, we need to
predict $m(x'_t|x_{<t})$ well for all $x'_t\in\X$. $m(|)$ also
converges off-sequence for $\bar x_t\neq x_t$ (to zero as it
should be), but the bound (third line) is much weaker than the
on-sequence bound (first line), so rapid convergence cannot be
concluded, unlike for $M$, where
$M(x_t|x_{<t})\stackrel{fast}\longrightarrow 1$ implies $M(\bar
x_t|x_{<t})\stackrel{fast}\longrightarrow 0$, since
$\sum_{x'_t}M(x'_t|x_{<t})\leq 1$. Consider an environment
$x_{1:\infty}$ describable in 500 bits, then bound $(vi)$ line 2
does not exclude $m(\bar x_t|x_{<t})$ from being 1 (maximally
wrong) for all $t=1..2^{500}$; with asymptotic convergence being
of pure academic interest.

$\neg(vi)$ The situation is provably worse in the probabilistic
case. There are computable measures $\mu$ for which neither
$m(x_t|x_{<t})$ nor $m_{norm}(x_t|x_{<t})$ converge to
$\mu(x_t|x_{<t})$ for any $x_{1:\infty}$.

$(vii)$ Since $(vi)$ implies $(vii)$ by continuity, we have
convergence of the instantaneous losses for computable
environments $x_{1:\infty}$, but since we do not know the speed of
convergence off-sequence, we do not know how fast the losses
converge to optimum.

$\neg(vii)$ Non-convergence $\neg(vi)$ does not necessarily imply
that $\Lambda_m$ is not self-optimizing, since different
predictive functions can lead to the same predictor $\Lambda$. But
it turns out that $\Lambda_m$ is not self-optimizing even in
Bernoulli environments $\mu$ for particular losses $\ell$ (first
line). A similar result holds for {\em any} {\em non-degenerate
loss function} (especially  for the error-loss, cf.\
(\ref{eqMDLk})), for specific choices of the universal
Turing-machine $U$ (second line). Loss $\ell$ is defined to be
non-degenerate {\em iff} $\bigcap_{x\in\X}\{\tilde
y\,:\,\ell_{x\tilde y}=\min_y\ell_{xy}\} = \{\}$. Assume the
contrary that a {\it single} action $\tilde y$ is optimal for {\it
every} outcome $x$, i.e.\ that ($\arg\min_y$ can be chosen such
that) $\arg\min_y\ell_{xy}=\tilde y\,\forall x$. This implies
$y_t^{\Lambda_\rho}=\tilde y\,\forall\rho$, which implies
$l_t^{\Lambda_m}/l_t^{\Lambda_\mu}\equiv 1$. So the non-degeneracy
assumption is necessary (and sufficient).

\section{Proof of Theorem \ref{thmDProp}}\label{secmProof}

\paranodot{(o)}
The first two properties are due to Levin and are proven in
\cite{Levin:73random} and \cite[Th.3.4]{Zvonkin:70}, respectively.
The third property is an easy corollary from G\'acs result
\cite{Gacs:83}, which says that if $g$ is some monotone
co-enumerable function for which $\Km(x)-\KM(x)\leq g(\l(x))$
holds for all $x$, then $g(n)$ must be $\geqa  K(n)$. Assume
$\Km(x)-\KM(x)\geq \lb\,\l(x)$ only for finitely many $x$ only.
Then there exists a $c$ such that $\Km(x)-\KM(x)\leq \lb\,\l(x)+c$
for {\em all} $x$. G\'acs' theorem now implies $\lb\,n+c\geqa
K(n)\,\forall n$, which is wrong due to Kraft's inequality $\sum_n
2^{-K(n)}\leq 1$.

\paranodot{(i)}
The first line is proven in \cite[Th.4.5.4]{Li:97}. Exponentiating
this result gives $m(x)\geq c_\mu\mu(x)\,\forall
x,\mu\in \M_{comp}^{msr}$, i.e.\
$m\geqm\M_{comp}^{msr}$. Exponentiation of $\neg(o)$
gives $m(x)\leq\MM(x)/\l(x)$, which implies
$m(x)\not\geqm\MM(x)\in\M_{enum}^{semi}$, i.e.\
$m\not\geqm\M_{enum}^{semi}$.

\paranodot{(ii)}
is obvious from the definition of $\Km$ and $m$. %

\paranodot{$\neg$(iii)}
Simple violation of the semimeasure property can be
inferred indirectly from $(i),(iv),\neg(vi)$ and Theorem
\ref{thPredRel}$b$. To prove $\neg(iii)$ we first
note that $\Km(x)<\infty$ for all finite strings $x\in\X^*$, which
implies $m(x_{1:n})>0$. Hence, whenever $\Km(x_{1:n})=\Km(x_{<n})$,
we have $\sum_{x_n}m(x_{1:n})> m(x_{1:n})=m(x_{<n})$, a violation
of the semimeasure property. $\neg(iii)$ now follows from
$
  \#\{t\leq n: \sum_{x_t}m(x_{1:t})\leq m(x_{<t})\}
  \leq \#\{t\leq n: \Km(x_{1:t})\neq \Km(x_{<t})\}
  \leq \sum_{t=1}^n \Km(x_{1:t})-\Km(x_{<t})
  = \Km(x_{1:n})
$, where we exploited $(ii)$ in the last inequality.

\paranodot{(iv)} immediate from $(ii)$.%

\paranodot{$\neg$(iv)}
(first line) follows from the fact that equality does not even
hold within an additive constant, i.e.\ $\Km(yx)\not\equa \Km(x|y)+\Km(y)$. The proof of the latter is
similar to the one for $K$ (see \cite{Li:97}). $\neg(iv)$
(second line) follows within $\log$ from $\Km=K+O(\log)$ and
$K(yx)=K(x|y)+K(y)+O(\log)$ \cite{Li:97}.

\paranodot{(v)} immediate from definition. %

\paranodot{(vi)}
$ \#\{t\leq n:m(x_t|x_{<t})\neq 1\}
  \leq \sum_{t=1}^n 2|1-m(x_t|x_{<t})|
  \leq -\sum_{t=1}^n \lb\,m(x_t|x_{<t})
  = -\lb\,m(x_{1:n})
  = \Km(x_{1:n})
  < \infty.
$
In the first inequality we used $m:=m(x_t|x_{<t})\in 2^{-\Set
N_0}$, hence $1\leq 2|1-m|$ for $m\neq 1$. In the second
inequality we used $1-m\leq-\odt\lb\,m$, valid for
$m\in[0,\odt]\cup\{1\}$. In the first equality we used (the $\log$
of) \mrcp\ $n$ times. For computable $x_{1:\infty}$ we have
$\sum_{t=1}^\infty|1-m(x_t|x_{<t})| \leq \odt\Km(x_{1:\infty}) <
\infty$, which implies $m(x_t|x_{<t})\to 0$ (fast if
$\Km(x_{1:\infty})$ is of reasonable size).
This shows the first two lines of $(vi)$.
The last line is shown as follows: Fix a sequence $x_{1:\infty}$ and
define $\Q:=\{x_{<t}\bar x_t \,:\, t\in\Set N,\,\bar x_t\neq
x_t\}$. $\Q$ is a prefix-free set of finite strings. For any such
$\Q$ and any semimeasure $\mu$, one can show that $\sum_{x\in
\Q}\mu(x)\leq 1$.\footnote{This follows from $1\geq\mu(A\cup B)\geq
\mu(A)+\mu(B)$ if $A\cap B=\{\}$, $\Gamma_x\cap\Gamma_y=\{\}$ if
$x$ not prefix of $y$ and $y$ not prefix of $x$, where
$\Gamma_x:=\{\omega:\omega_{1:\l(x)}=x\}$, hence $\sum_{x\in
\Q}\mu(\Gamma_x)\leq\mu(\bigcup_{x\in \Q}\Gamma_x)\leq 1$, and
noting that $\mu(x)$ is actually an abbreviation for
$\mu(\Gamma_x)$.} Since $M$ is a semimeasure lower bounded by $m$
we get
\beqn
  \sum_{t=1}^n m(x_{<t}\bar x_t)
  \;\leq\; \sum_{t=1}^\infty m(x_{<t}\bar x_t)
  \;=\; \sum_{x\in \Q} m(x)
  \;\leq\; \sum_{x\in \Q} M(x)
  \;\leq\; 1.
\eeqn
With this, and using monotonicity of $m$ we get
\beqn
  \sum_{t=1}^n m(\bar x_t|x_{<t})
  = \sum_{t=1}^n {m(x_{<t}\bar x_t)\over
  m(x_{<t})}
  \leq \sum_{t=1}^n {m(x_{<t}\bar x_t)\over
  m(x_{1:n})}
  \leq {1\over m(x_{1:n})}
  = 2^{\Km(x_{1:n})}
\eeqn
Finally, for an infinite sum to be finite, its elements must
converge to zero.

\paranodot{$\neg$(vi)}
follows from the non-denseness of the range of $m_{(norm)}$:
We choose $\mu(1|x_{<t})={3\over 8}$, hence $\mu(0|x_{<t})={5\over
8}$. Since $m(x_t|x_{<t})\in 2^{-\Set N_0}=\{1,\odt,\odf,{1\over
8},...\}$, we have $|m(x_t|x_{<t})-\mu(x_t|x_{<t})|\geq{1\over
8}\,\forall t,\,\forall x_{1:\infty}$. Similarly for
\beqn\textstyle
  m_{norm}(x_t|x_{<t}) \;=\;
  {m(x_t|x_{<t})\over m(0|x_{<t})+m(1|x_{<t})}
  \;\in\; \{ {2^{-n}\over 2^{-n}+2^{-m}}:n,m\!\in\!\Set N_0 \} \;=
\eeqn\vspace{-2ex}
\beqn\textstyle
    =\; \{ {1\over 1+2^z}:z\!\in\!\Set Z\}
  \;=\; {1\over 1+2^{\Set Z}}
  \;=\; \{ ...,{1\over 9},{1\over 5},{1\over 3},\odt,
       {2\over 3},{4\over 5},{8\over 9},... \}
\eeqn
we choose $\mu(1|x_{<t})=1-\mu(0|x_{<t})={5\over 12}$, which
implies $|m_{norm}(x_t|x_{<t})-\mu(x_t|x_{<t})|\geq{1\over
12}$ $\forall t$, $\forall x_{1:\infty}$.

\paranodot{(vii)}
The first line follows from $(vi)$ and
Theorem \ref{thPredRel}d. That normalization does not affect the
predictor, follows from the definition of $y_t^{\Lambda_\rho}$
(\ref{xlrdef}) and the fact that $\arg\min()$ is not affected by
scaling its argument.

\paranodot{$\neg$(vii)}
Non-convergence of $m$ does not necessarily imply non-convergence
of the losses. For instance, for $\Y=\{0,1\}$, and
$\omega'_t:=1/0$ for
$\mu(1|x_{<t}){>\atop<}\gamma := {\ell_{01}-\ell_{00}\over
\ell_{01}-\ell_{00}+\ell_{10}-\ell_{11}}$, one can show
that $y_t^{\Lambda_\mu}=y_t^{\Lambda_{\omega'}}$, hence
convergence of $m(x_t|x_{<t})$ to 0/1 and not to $\mu(x_t|x_{<t})$
could nevertheless lead to correct predictions.

Consider now $y\in\Y=\{0,1,2\}$. To prove the first line of
$\neg$(vii) we define a loss function such that
$y_t^{\Lambda_\mu}\neq y_t^{\Lambda_\rho}$ for any $\rho$ with
same range as $m_{norm}$ and for some $\mu$. The loss function
$\ell_{x0}=x$, $\ell_{x1}={3\over 8}$, $\ell_{x2}={2\over
3}(1-x)$, and $\mu:=\mu(1|x_{<t})={2\over 5}$ will do. The
$\rho$-expected loss under action $y$ is $l_\rho^y:=\sum_{x_t=0}^1
\rho(x_t|x_{<t})\ell_{x_t y}$; $l_\rho^0=\rho$, $l_\rho^1={3\over
8}$, $l_\rho^2={2\over 3}(1-\rho)$ with $\rho:=\rho(1|x_{<t})$
(see Figure).
\def\spfrac#1#2{{{#1\over #2}}}
\def\figlossdef{
\begin{picture}(200,150)(-50,-10)
\put(0,0){\vector(1,0){140}}\put(130,-2){\makebox(0,0)[lt]{$\rho$}}
\put(-1,120){\line(1,0){2}}\put(-2,120){\makebox(0,0)[rc]{$1$}}
\put(0,0){\vector(0,1){135}}\put(-2,125){\makebox(0,0)[rb]{$l_\rho^y$}}
\put(120,-1){\line(0,1){2}}\put(120,-2){\makebox(0,0)[ct]{$1$}}
\put(0,0){\dashbox(120,120){}}
\put(1.85,0){\circle*{1.3}}
\put(3.64,0){\circle*{1.5}}
\put(7.06,0){\circle*{3}}
\put(13.33,0){\circle*{4}}
\put(24,0){\circle*{4}}\put(24,-2){\makebox(0,0)[ct]{$\spfrac{1}{5}$}}
\put(40,0){\circle*{4}}\put(40,-2){\makebox(0,0)[ct]{$\spfrac{1}{3}$}}
\put(60,0){\circle*{4}}\put(60,-2){\makebox(0,0)[ct]{$\spfrac{1}{2}$}}
\put(80,0){\circle*{4}}\put(80,-2){\makebox(0,0)[ct]{$\spfrac{2}{3}$}}
\put(96,0){\circle*{4}}\put(96,-2){\makebox(0,0)[ct]{$\spfrac{4}{5}$}}
\put(106.67,0){\circle*{4}}
\put(112.94,0){\circle*{3}}
\put(116.36,0){\circle*{1.5}}
\put(118.15,0){\circle*{1.3}}
\put(0,0){\line(1,1){120}}
\put(-2,0){\makebox(0,0)[rc]{$\ell_{00}=0$}}
\put(119,120){\line(1,0){2}}\put(122,120){\makebox(0,0)[lc]{$1=\ell_{10}$}}
\put(100,100){\makebox(0,0)[lt]{$l_\rho^0$}}
\put(0,45){\line(1,0){120}}
\put(-1,45){\line(1,0){2}}\put(-2,45){\makebox(0,0)[rc]{$\ell_{01}=\,^3\!\!/\!_8$}}
\put(119,45){\line(1,0){2}}\put(122,45){\makebox(0,0)[lc]{$^3\!\!/\!_8=\ell_{11}$}}
\put(100,47){\makebox(0,0)[lb]{$l_\rho^1$}}
\put(120,0){\line(-3,2){120}}
\put(-1,80){\line(1,0){2}}\put(-2,80){\makebox(0,0)[rc]{$\ell_{02}=\,^2\!\!/\!_3$}}
\put(119,0){\line(1,0){2}}\put(122,2){\makebox(0,0)[lb]{$0=\ell_{12}$}}
\put(100,13){\makebox(0,0)[lb]{$l_\rho^2$}}
\put(40,0){\dashbox(20,120){}}
\put(0,40){\dashbox(60,0){}}
\put(-1,40){\line(1,0){2}}
\put(-2,39){\makebox(0,0)[rt]{$^1\!\!/\!_3$}}
\put(0,0){\dashbox(48,48){}}
\put(-1,48){\line(1,0){2}}
\put(-2,50){\makebox(0,0)[rb]{$^2\!\!/\!_5$}}
\put(48,-1){\line(0,1){2}}
\put(48,-2){\makebox(0,0)[ct]{$\spfrac{2}{5}$}}
\end{picture}
}
\begin{wrapfigure}{r}{45mm}
\scriptsize\unitlength=0.2mm
\thinlines\figlossdef
\end{wrapfigure}
Since $l_\mu^0=l_\mu^2={2\over 5}>{3\over 8}=l_\mu^1$, we have
$y_t^{\Lambda_\mu}=1$ and $l_t^{\Lambda_\mu}=l_\mu^1={3\over 8}$.
For $\rho\leq{1\over 3}$, we have $l_\rho^0<l_\rho^1<l_\rho^2$,
hence $y_t^{\Lambda_\rho}=0$ and
$l_t^{\Lambda_\rho}=l_\mu^0={2\over 5}$. For $\rho\geq\odt$, we
have $l_\rho^2<l_\rho^1<l_\rho^0$, hence $y_t^{\Lambda_\rho}=2$
and $l_t^{\Lambda_\rho}=l_\mu^2={2\over 5}$. Since
$m_{norm}\not\in({1\over 3},\odt)$, $\Lambda_{m_{norm}}$ predicts
$0$ or $2$, hence $l_t^{\Lambda_m}=l_\mu^{0/2}={2\over 5}$. Since
$\Lambda_{m_{norm}}=\Lambda_m$, this shows that
$l_t^{\Lambda_m}/l_t^{\Lambda_\mu}={16\over 15}>1$. The constant
${16\over 15}$ can be enlarged to ${6\over 5}-\eps$ by setting
$\ell_{x1}={1\over 3}+\eps$ instead of ${3\over 8}$.

For $\Y=\{0,...,|\Y|-1\}$, $|\Y|>3$, we extend the loss function
by defining $\ell_{xy}=1$ $\forall y\geq 3$, ensuring that actions
$y\geq 2$ are never favored. With this extension, the analysis of
the $|\Y|=3$ case applies, which finally shows $\neg(vii)$. In
general, a non-dense range of $\rho(x_t|x_{<t})$ implies
$l_t^{\Lambda_\rho}\not\to l_t^{\Lambda_\mu}$, provided $|\Y|\geq
3$.

We now construct a monotone universal Turing machine $U$
satisfying $\neg(vii)$ (second line).
In case where
ambiguities in the choice of $y$ in $\arg\min_y\ell_{xy}$ matter
we consider the set of solutions $\{\arg\min_y\ell_{xy}\}:=\{\tilde
y:\ell_{x\tilde y}=\min_y\ell_{xy}\}\neq\{\}$.
We define a one-to-one (onto $A$) decoding function $d:\B^s\to A$
with $A=\{0^{s+1}\}\cup 1\B^s\setminus 1\{0^s\}\subset\X^{s+1}$ as
$d(0_{1:s})=0_{1:s+1}$ and $d(x_{1:s})=1x_{1:s}$ for $x_{1:s}\neq
0_{1:s}$ with a large $s\in\Set N$ to be determined later. We
extend $d$ to $d:(\B^s)^*\to A^*$ by defining
$d(z_1...z_k)=d(z_1)...d(z_k)$ for $z_i\in\B^s$ and define the
inverse coding function $c:A\to\B^s$ and its extension
$c:A^*\to(\B^s)^*$ by $c=d^{-1}$.
Roughly, $U$
is defined as $U(1p_{1:sn}0_{1:s})=d(p_{1:sn})0_{1:s+1}$. More
precisely, if the first bit of the binary input tape of $U$ contains 1,
$U$ decodes the successive blocks of size $s$, but always
withholds the output until a block $0_{1:s}$ appears. $U$ is
obviously monotone. Universality will be guaranteed by defining
$U(0p)$ appropriately, but for the moment we set $U(0p)=\epstr$.
It is easy to see that for $x\in A^*$ we have
\beq\label{eqEUKm}
\begin{array}{rcl}
\Km(x0) =& \Km(x 0_{1:s+1})            &=\;\l(c(x))+s+1 \qmbox{and} \\
\Km(x1) =& \Km(x1z0_{1:s+1}) &=\;\l(c(x))+2s+1,
\end{array}
\eeq
where $z$ is any string
of length $s$. Hence, $m_{norm}(0|x)=[1+2^{-s}]^{-1}\toinfty{s} 1$
and $m_{norm}(1|x)=[1+2^s]^{-1}\toinfty{s} 0$. For $t-1\in(s+1)\Set
N$ we get $l_m^{y_t}:=\sum_{x_t}m_{norm}(x_t|x_{<t})
\ell_{x_t y_t}\toinfty{s}\ell_{0 y_t}$. This implies
\beq\label{eqEUm}
  y_t^{\Lambda_m}\in\{\arg\min_{y_t}l_m^{y_t}\}
  \subseteq \{\arg\min_y\ell_{0y}\}
  \quad\mbox{for sufficiently large finite $s$}.
\eeq
We now define $\mu(z)=|A|^{-1}=2^{-s}$ for $z\in A$ and $\mu(z)=0$
for $z\in\X^{s+1}\setminus A$, extend it to $\mu(z_1...z_k) :=
\mu(z_1)\cdot...\cdot\mu(z_k)$ for $z_i\in\X^{s+1}$, and finally
extend it uniquely to a measure on $\X^*$ by
$\mu(x_{<t}):=\sum_{x_{t:n}}\mu(x_{1:n})$ for $\Set N\ni t\leq
n\in(s+1)\Set N$. For $x\in A^*$ we have
$\mu(0|x)=\mu(0)=\mu(0_{1:s+1})=2^{-s}\toinfty{s} 0$ and
$\mu(1|x)=\mu(1)=\sum_{y\in\X^s}\mu(1y)=\sum_{z\in
A\setminus\{0^{s+1}\}}\mu(z) =(2^s-1)\cdot
2^{-s}=1-2^{-s}\toinfty{s} 1$. For $t-1\in(s+1)\Set N$ we get
$l_\mu^{y_t}:=\sum_{x_t}\mu(x_t|x_{<t}) \ell_{x_t
y_t}\toinfty{s}\ell_{1 y_t}$. This implies
\beq\label{eqEUmu}
  y_t^{\Lambda_\mu}\in\{\arg\min_{y_t}l_\mu^{y_t}\}
  \subseteq \{\arg\min_y\ell_{1y}\}
  \quad\mbox{for sufficiently large finite $s$}.
\eeq
By definition, $\ell$ is non-degenerate iff
$\{\arg\min_y\ell_{0y}\}\cap\{\arg\min_y\ell_{1y}\}=\{\}$. This,
together with (\ref{eqEUm}) and (\ref{eqEUmu}) implies
$y_t^{\Lambda_m}\neq y_t^{\Lambda_\mu}$, which implies
$l_t^{\Lambda_m}\neq l_t^{\Lambda_\mu}$ (otherwise the choice
$y_t^{\Lambda_m}= y_t^{\Lambda_\mu}$ would have been possible),
which implies
$l_t^{\Lambda_m}/l_t^{\Lambda_\mu}=c>1$ for $t-1\in(s+1)\Set
N$, i.e.\ for infinitely many $t$.

What remains to do is to extend $U$ to a universal Turing machine.
We extend $U$ by defining $U(0zp)=U'(p)$ for any $z\in\B^{3s}$,
where $U'$ is some universal Turing machine.
Clearly, $U$ is now universal. We have to show that this extension
does not spoil the preceding consideration, i.e.\ that the shortest
code of $x$ has sufficiently often the form $1p$ and sufficiently
seldom the form $0p$. Above, $\mu$ has been chosen in such a way that
$c(x)$ is a Shannon-Fano code for $\mu$-distributed strings,
i.e.\ $c(x)$ is with high $\mu$-probability a shortest code of $x$.
More precisely, $\l(c(x))\leq\Km_T(x)+s$ with $\mu$-probability at
least $1-2^{-s}$, where $\Km_T$ is the monotone complexity w.r.t.\
any decoder $T$, especially $T=U'$. This implies
$
  \min_p\{\l(0p) : U(0p)=x*\}
  = 3s+1+\Km_{U'}(x)
  \geq 3s+1+\l(c(x))-s
  > \l(c(x))+s+1
  \geq \min_p\{\l(1p) : U(1p)=x*\}, $ %
where the first $\geq$ holds
with high probability ($1-2^{-s}$). This shows that the
expressions (\ref{eqEUKm}) for $\Km$ are with high probability not
affected by the extension of $U$. Altogether this shows
${l_t^{\Lambda_m}/l_t^{\Lambda_\mu}} \;\;\not\!\!\!\toinfty{t} 1$
with high probability.
\qed

\section{Outlook and Open Problems}\label{secOpen}

\paragraph{Speed of off-sequence convergence of $m$ for computable environments}
The probably most interesting open question is how fast $m(\bar
x_t|x_{<t})$ converges to zero in the deterministic case.

\paragraph{Non-self-optimizingness for general $U$ and $\ell$}
Another open problem is whether for every non-degenerate
loss-function, self-optimizingness of $\Lambda_m$ can be violated.
We have shown that this is the case for particular choices of
the universal Turing machine $U$. If $\Lambda_m$
were self-optimizing for some $U$ and general loss, this would be an
unusual situation in Algorithmic Information Theory, where
properties typically hold for all or no $U$.
So we expect $\Lambda_m$ not to be self-optimizing for general
loss and $U$ (particular $\mu$ of course).
A first step may be to try to prove that for all $U$ there exists
a computable sequence $x_{1:\infty}$ such that $K_U(x_{<t}\bar
x_t)<K_U(x_{<t}x_t)$ for infinitely many $t$
(which shows $\neg(vii)$ for $K$ and error-loss), and then try to
generalize to probabilistic $\mu$, $\Km$, and general loss
functions.

\paragraph{Other complexity measures}
This work analyzed the predictive properties of the monotone
complexity $\Km$. This choice was motivated by the fact that $m$ is
the MDL approximation of the sum $M$, and $\Km$ is {\em very} close
to $\KM$. We expect all other (reasonable) alternative
complexity measure to perform worse than $\Km$. But we
should be careful with precipitative conclusions, since closeness
of unconditional predictive functions not necessarily implies good
prediction performance, so distantness may not necessarily imply
poor performance.
What is easy to see is that $K(x)$ (and $K(x|\l(x))$) are completely
unsuitable for prediction, since $K(x0)\equa K(x1)$ (and $K(x0|\l(x0))\equa
K(x1|\l(x1))$), which implies that the predictive functions do not
even converge for deterministic computable environments.
Note that the larger a semimeasures, the more distributions it
dominates, the better its predictive properties. This simple rule
does not hold for non-semimeasures. Although $M$ predicts better
than $m$ predicts better than $k$ in accordance with
(\ref{Krels}), $2^{-K(x|\l(x))}\geqm\MM(x)$ is a bad predictor
disaccording with (\ref{Krels}).
Besides the discussed
prefix Kolmogorov complexity $K$, %
monotone complexity $\Km$, and %
Solomonoff's universal prior $M=2^{-\KM}$, %
one may investigate the predictive properties of the historically first %
plain Kolmogorov complexity $C$, %
Schnorr's process complexity, %
Chaitin's complexity $Kc$, %
Cover's extension semimeasure $Mc$, %
Loveland's uniform complexity, %
Schmidhuber's cumulative $K^E$ and general $K^G$ complexity and corresponding measures, %
Vovk's predictive complexity $K\!P$, %
Schmidhuber's speed prior $S$, %
Levin complexity $Kt$, %
and several others \cite{Li:97,Vovk:98,Schmidhuber:00toe}.
Many properties and relations are known for the unconditional
versions, but little relevant for prediction of the conditional
versions is known.

\paragraph{Two-part MDL}
We have approximated $M(x):=\sum_{p:U(p)=x*}2^{-\l(p)}$ by its
dominant contribution $m(x)=2^{-\Km(x)}$, which we have
interpreted as deterministic or one-part universal MDL. There is
another representation of $M$ due to Levin \cite{Zvonkin:70} as a
mixture over semi-measures:
$M(x)=\sum_{\nu\in\M_{enum}^{semi}}2^{-K(\nu)}\nu(x)$ with
dominant contribution $m_2(x)=2^{-\Km_2(x)}$ and universal
two-part MDL
$\Km_2(x):=\min_{\nu\in\M_{enum}^{semi}}\{-\lb\,\nu(x)+K(\nu)\}$.
MDL ``lives'' from the validity of this approximation. $K(\nu)$ is
the complexity of the probabilistic model $\nu$, and
$-\lb\,\nu(x)$ is the (Shannon-Fano) description length of data
$x$ in model $\nu$. MDL usually refers to two-part MDL, and not to
one-part MDL. A natural question is to ask about the predictive
properties of $m_2$, similarly to $m$. $m_2$ is even closer to $M$
than $m$ is ($m_2\eqm M$), but is also not a semi-measure. Drawing
the analogy to $m$ further, we conjecture slow posterior
convergence $m_2\to\mu$ w.p.1 for computable probabilistic
environments $\mu$. In \cite{Barron:91}, MDL has been shown to
converge for computable i.i.d.\ environments.

\paranodot{More abstract proofs}
showing that violation of some of the criteria $(i)-(iv)$
necessarily lead to violation of $(vi)$ or $(vii)$ may deal with a
number of complexity measures simultaneously. For instance, we
have seen that any non-dense posterior set $\{\tilde
k(x_t|x_{<t})\}$ implies non-convergence and
non-self-optimizingness; the particular structure of $m$ did not
matter.

\paragraph{Extra conditions}
Non-convergence or non-self-optimizingness of $m$ do not
necessarily mean that $m$ fails in practice. Often one knows more
than that the environment is (probabilistically) computable, or
the environment possess certain additional properties, even if
unknown. So one should find sufficient and/or necessary extra
conditions on $\mu$ under which $m$ converges / $\Lambda_m$
self-optimizes rapidly. The results of this work have shown that
for $m$-based prediction one {\em has} to make extra assumptions
(as compared to $\MM$). It would be interesting to characterize
the class of environments for which universal MDL alias $m$ is a
good predictive approximation to $M$. Deterministic computable
environments were such a class, but a rather small one, and
convergence is possibly slow.



{\small
\begin{thebibliography}{Hut01b}

\bibitem[BC91]{Barron:91}
A.~R. Barron and T.~M. Cover.
\newblock Minimum complexity density estimation.
\newblock {\em IEEE Transactions on Information Theory}, 37:1034--1054, 1991.

\bibitem[G{\'a}c83]{Gacs:83}
P.~G{\'a}cs.
\newblock On the relation between descriptional complexity and algorithmic
  probability.
\newblock {\em Theoretical Computer Science}, 22:71--93, 1983.

\bibitem[Hut01a]{Hutter:01alpha}
M.~Hutter.
\newblock Convergence and error bounds of universal prediction for general
  alphabet.
\newblock {\em Proceedings of the 12th Eurpean Conference on Machine Learning
  (ECML-2001)}, pages 239--250, 2001.

\bibitem[Hut01b]{Hutter:99errbnd}
M.~Hutter.
\newblock New error bounds for {Solomonoff} prediction.
\newblock {\em Journal of Computer and System Sciences}, 62(4):653--667, 2001.

\bibitem[Hut02]{Hutter:02spupper}
M.~Hutter.
\newblock Convergence and loss bounds for {Bayesian} sequence prediction.
\newblock Technical Report IDSIA-09-01, IDSIA, Manno(Lugano), CH, 2002.
\newblock http://arxiv.org/abs/cs.LG/0301014.

\bibitem[KV86]{Kumar:86}
P.~R. Kumar and P.~P. Varaiya.
\newblock {\em Stochastic Systems: Estimation, Identification, and Adaptive
  Control}.
\newblock Prentice Hall, Englewood Cliffs, NJ, 1986.

\bibitem[Lev73]{Levin:73random}
L.~A. Levin.
\newblock On the notion of a random sequence.
\newblock {\em Soviet Math. Dokl.}, 14(5):1413--1416, 1973.

\bibitem[LV97]{Li:97}
M.~Li and P.~M.~B. Vit\'anyi.
\newblock {\em An introduction to {Kolmogorov} complexity and its
  applications}.
\newblock Springer, 2nd edition, 1997.

\bibitem[Sch00]{Schmidhuber:00toe}
J.~Schmidhuber.
\newblock Algorithmic theories of everything.
\newblock Report IDSIA-20-00, quant-ph/0011122, {IDSIA}, Manno (Lugano),
  Switzerland, 2000.

\bibitem[Sol64]{Solomonoff:64}
R.~J. Solomonoff.
\newblock A formal theory of inductive inference: Part 1 and 2.
\newblock {\em Inform. Control}, 7:1--22, 224--254, 1964.

\bibitem[Sol78]{Solomonoff:78}
R.~J. Solomonoff.
\newblock Complexity-based induction systems: comparisons and convergence
  theorems.
\newblock {\em IEEE Trans. Inform. Theory}, IT-24:422--432, 1978.

\bibitem[VW98]{Vovk:98}
V.~G. Vovk and C.~Watkins.
\newblock Universal portfolio selection.
\newblock In {\em Proceedings of the 11th Annual Conference on Computational
  Learning Theory ({COLT}-98)}, pages 12--23, New York, 1998. ACM Press.

\bibitem[ZL70]{Zvonkin:70}
A.~K. Zvonkin and L.~A. Levin.
\newblock The complexity of finite objects and the development of the concepts
  of information and randomness by means of the theory of algorithms.
\newblock {\em Russian Mathematical Surveys}, 25(6):83--124, 1970.

\end{thebibliography}
}
\end{document}